\documentclass[letterpaper, 10 pt, conference]{ieeeconf}  
\usepackage[utf8]{inputenc}
\usepackage{algorithm}
\usepackage{algpseudocode}
\usepackage{cite}
\usepackage{hyperref}
\usepackage{graphicx} 
\usepackage{amsmath} 
\usepackage{amssymb}  
\usepackage{subfig}
\usepackage{xcolor}
\usepackage{tabularx}
\usepackage[T1]{fontenc}

\IEEEoverridecommandlockouts                              

\title{\LARGE \bf
Accelerating Laboratory Automation Through Robot Skill Learning For Sample Scraping*
}

\author{Gabriella Pizzuto$^{1}$, Hetong Wang$^{2}$, Hatem Fakhruldeen$^{3}$, Bei Peng$^{1}$, Kevin S. Luck$^4$ and Andrew I. Cooper$^{3}$
\thanks{*This work was supported by the Leverhulme Trust through the Leverhulme Research Centre for Functional Materials Design, the H2020 ERC Synergy Grant Autonomous Discovery of Advanced Materials under grant agreement no. 856405 and the Royal Academy of Engineering under the Research Fellowship Scheme.}
\thanks{$^{1}$Gabriella Pizzuto and Bei Peng are with the Department of Computer Science, University of Liverpool, United Kingdom. 
{\newline Corresponding author: \textit{gabriella.pizzuto@liverpool.ac.uk}}}%
\thanks{$^{2}$Hetong Wang is with the School of Informatics, University of Edinburgh, United Kingdom.
}
\thanks{$^{3}$Hatem Fakhruldeen and Andrew I. Cooper are with the Department of Chemistry, University of Liverpool, United Kingdom.
}
\thanks{$^{4}$Kevin Sebastian Luck is with the Faculty of Science, Vrije Universiteit Amsterdam, Netherlands.}
}

\begin{document}

\maketitle
\thispagestyle{empty}
\pagestyle{empty}

\begin{abstract}
The use of laboratory robotics for autonomous experiments offers an attractive route to alleviate scientists from tedious tasks while accelerating material discovery for topical issues such as climate change and pharmaceuticals.
While some experimental workflows can already benefit from automation,
sample preparation is still carried out manually due to the high level of motor function and dexterity required when dealing with different tools, chemicals, and glassware.
A fundamental workflow in chemical fields is crystallisation, where one application is polymorph screening, i.e., obtaining a three dimensional molecular structure from a crystal. 
For this process, it is of utmost importance to recover as much of the sample as possible since synthesising molecules is both costly in time and money. 
To this aim, chemists scrape vials to retrieve sample contents prior to imaging plate transfer.
Automating this process is challenging as it goes beyond robotic insertion tasks due to a fundamental requirement of having to execute fine-granular movements within a constrained environment (sample vial).
Motivated by how human chemists carry out this process of scraping powder from vials, our work proposes a model-free reinforcement learning method for learning a scraping policy, leading to a fully autonomous sample scraping procedure. 
We first create a scenario-specific simulation environment with a Panda Franka Emika robot using a laboratory scraper that is inserted into a simulated vial, to demonstrate how a scraping policy can be learned successfully in simulation. 
We then train and evaluate our method on a real robotic manipulator in laboratory settings, and show that our method can autonomously scrape powder across various setups.

\end{abstract}
\section{Introduction}

Transforming materials discovery plays a pivotal role in addressing global challenges, for example, renewable and clean energy storage, sustainable polymers, drugs and therapeutics and packaging for consumer products towards a more circular economy.
To accelerate the frequency of fundamental breakthroughs in this field, human scientists can be supported by 'robot scientists' to carry out long-term experiments that are key to obtaining new materials.
The goal here is that robots will carry out tedious cognitive and manual tasks, whereas humans conduct experiments that are safe and mentally stimulating.
While there has been remarkable progress to introduce robotics in the field of laboratory automation~\cite{Thurow2022},~\cite{Fakhruldeen2022},~\cite{Lim2021},~\cite{Tom2024}, there remains an open gap in having robots carry out tasks beyond pre-programmed manipulation.
This hinders progress in developing autonomous robots that are capable of operating end-to-end in human labs, or at the very least continue scientists' experiments after hours and are able to support novel experimental protocols.

\begin{figure}[!tbp]
  \centering
  \includegraphics[width=0.49\textwidth]{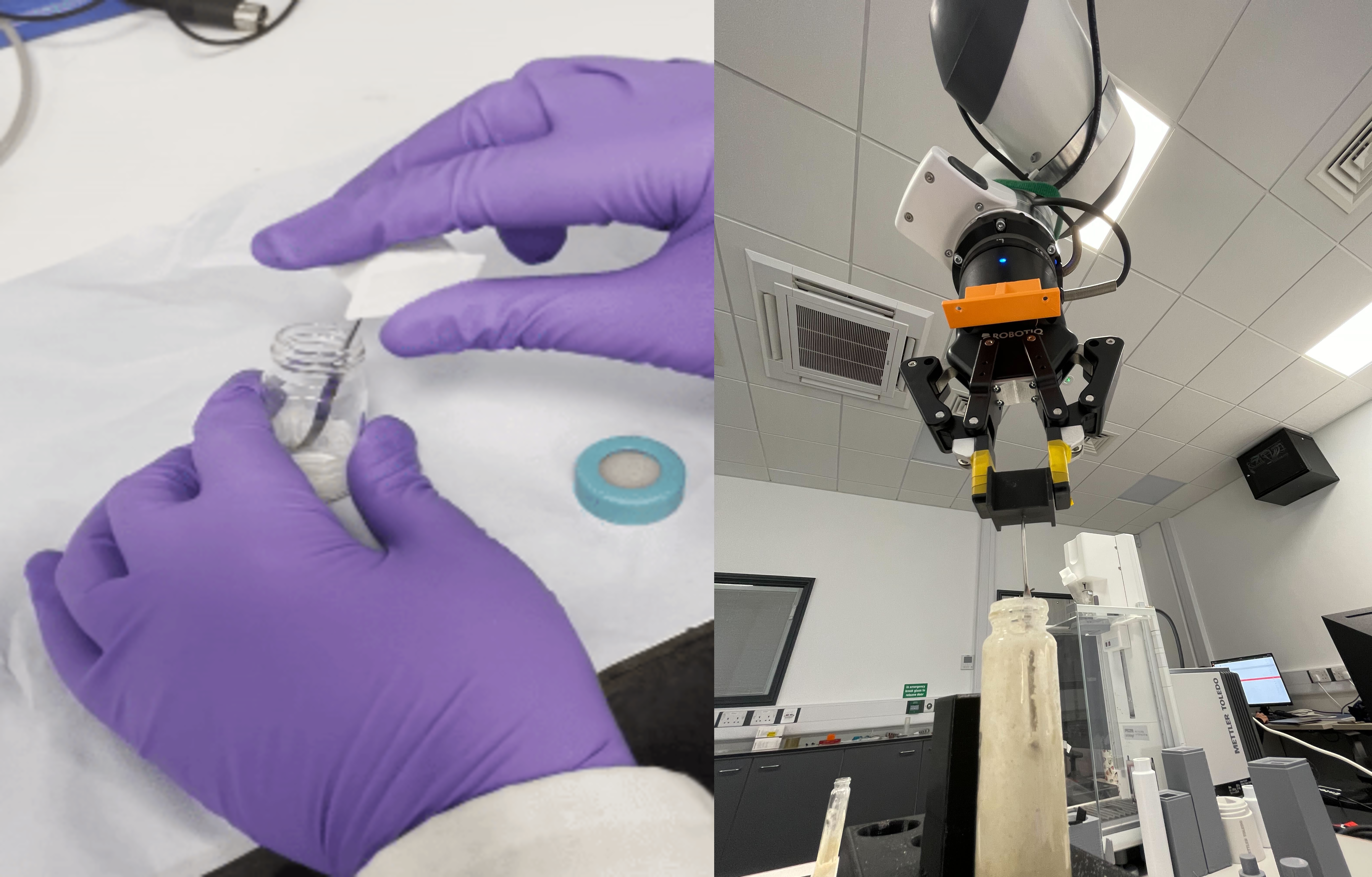}
  \caption{Overview of laboratory sample scraping, where (left) a human scientist scrapes crystals that generally form on vial walls by transcending along the walls vertically and (right) our robot learning to scrape via RL.}
  \label{fig:overview_scraping_task}
\end{figure}

To date, most laboratory robotics rely on pre-defined point-to-point motions without any sensory feedback or learning-based methods that would increase their adaptability to differing environmental conditions.
Learning would help to tackle the problem of environmental variation - learning algorithms are capable of adapting to variations in lighting conditions, displacements of material, etc.
Conversely, humans are able to use different modalities and carry out skills that gradually get better with experience.
This is not, and will never be, attainable with traditional laboratory automation due to the required dexterity and coordination between visual inputs and motor response that is trivial in humans~\cite{Seifrid2022}, ~\cite{Maffettone2023}.
In turn, robots were able to acquire a variety of practical skills that previously were not possible~\cite{Zhao22}.
Solving dexterous manipulation in laboratory tasks is an important problem as it enables a wide range of useful robotic workflows that otherwise are not possible. 
Consider the sample scraping problem as illustrated in Fig.~\ref{fig:overview_scraping_task}.
A human chemist would scrape the vial walls by translating vertically in a circular fashion.
Solving such tasks with traditional laboratory automation method is not possible due to the difficulty of requiring force feedback as the walls are scraped and different crystal morphology would need an adaptive method of scraping. 

In this work, we investigate the use of model-free reinforcement learning (RL) for use in laboratory automation to overcome previous challenges in skill acquisition for material discovery experimental workflows. 
We learn policies for the full laboratory task of scraping, which map proprioceptive and force observations to end effector displacement actions of a robotic arm. 
Moreover, we explore curriculum learning  ~\cite{Bengio2009},~\cite{Narvekar2020}, where we progressively increase the difficulty of the task through extending the size of the workspace as the training performance of our agent improves, to reduce training costs while increasing the overall task success. 
By doing so, we could successfully learn the more challenging insertion and scraping task, while the RL method without a curriculum fails to learn.
Finally, while training and evaluation of the scraping policy is performed in simulation, we report the required adjustments and findings from migrating these policies to a real robotic platform. 

In summary, the contributions of this work are:
\begin{itemize}
    \item A closed-loop formulation of an in-tube scraping task, a sub-task of a crystallisation workflow.
    \item A new benchmark task for robot learning in labs\footnote{A link to the benchmark will be added here upon publication.}: closed-space manipulation and extraction of material.
    \item A case study of applying deep reinforcement learning to a challenging motor skill (in-tube scraping) that is not possible with traditional laboratory automation.
    \item Exploiting curriculum learning to improve the learning performance of the RL agent in a more challenging insertion and scraping task.
    \item A demonstration of the learned robot behaviour both in simulation and in a real-world Chemistry lab.
\end{itemize}

To the best of our knowledge, this is the first research that uses learning-based methods for acquiring human-inspired motor skills for a crystallisation workflow in a real-world Chemistry laboratory.
Evaluations of the method with an experimental setup using tools and glassware from a real-world material discovery lab are aimed at demonstrating its applicability to real laboratory environments.


\section{Related Work}

\subsection{Laboratory Automation for Manipulation Skills}

Rapid progress in robotics, automation, and algorithmic efficiency has made it possible for autonomous robotic platforms to be used in laboratory environments.
For example, robotic manipulators have been used to measure solubility in multiple solvents~\cite{Shiri2021},~\cite{Pizzuto2022}, and for carrying out a Michael reaction~\cite{Lim2021}, while mobile manipulators have exceeded human-level ability to search for improved photocatalysts for hydrogen production from water~\cite{Burger2020}. 
However, whilst these are excellent endeavours in removing these tedious and repetitive tasks from the scientists' daily routines, the robotic platforms do not exhibit the same skill as a human chemist would. 
In fact, the robotic skills in these experiments were pre-programmed and carried out without sensory feedback~\cite{Jiang2023}.
When sensory feedback was introduced, these were also limited to pick-and-place of vials and racks~\cite{Butterworth2023}, rather than irregular objects such as laboratory tools.
In practice, usage of robotic platforms as such would limit deployment of autonomous robots for a vast array of experiments as they would not be capable of generalising to new reactions, or would necessitate oversimplification of workflows leading to poorer results.

The automation of laboratories for life sciences has stricter protocols when compared to other fields, and as a result several works have demonstrated the usage of dual-arm robots to mimic human way of carrying out laboratory procedures.
It was demonstrated how a dual-arm robotic manipulator can be used for sample preparation and measurement in ~\cite{Fleischer2021},~\cite{Fleischer2016} and ~\cite{Joshi2019}.
Here, the authors focused on automating the instrument to facilitate the task of using laboratory instruments such as pipettes.
Conversely, we concentrate our research endeavours on having the robotic manipulator learn the task through goal-conditioned reinforcement learning which allows us to use laboratory tools that do not have to be automated.
This approach also enables us to address laboratory tasks that require human adaptability when dealing with different glassware and molecules that have distinct chemical properties such as hygroscopy, cohesiveness, hardness, amongst other factors.

\subsection{Reinforcement Learning for Contact-rich Manipulation}
There have been several methods proposed to accelerate automation of contact-rich robotic skill learning, primarily for insertion tasks such as peg-in-hole assembly. 
For example, deep reinforcement learning-based assembly search strategies to align the peg with the hole~\cite{Zhang21},~\cite{Yasutomi21}, industrial insertion tasks with tighter clearance~\cite{Inoue17} or using meta-reinforcement learning for generalising across a variety of different insertion tasks~\cite{Zhao22}.
Our task in this paper is similar to robotic insertion; however, it goes beyond any insertion task as the tool needs to maintain contact with the walls of the cylindrical vessel, which is also reflected through the reward function, similar to the constraint within the learning objective in~\cite{Pizzuto21}.

While deep reinforcement learning (DRL)-based policies are increasingly being used and shown to have the potential to solve assembly skills in real-world applications, there still exists an open gap in having such methods for real laboratory tasks.
One potential reason for this could be that laboratory automation has traditionally been done in open-loop, which hinders the interaction of the agent obtaining feedback from its environment.
We aim to contribute to this area by proposing a novel way of laboratory skill learning using goal-conditioned reinforcement learning~\cite{Liu22} for a task that has not been addressed before, while demonstrating the important role that DRL solutions can have when adopted in real-world laboratories to solve tasks that are unattainable with conventional methods.

\section{Methodology}

\begin{figure*}[]
    \centering
    \includegraphics[width=0.95\textwidth]{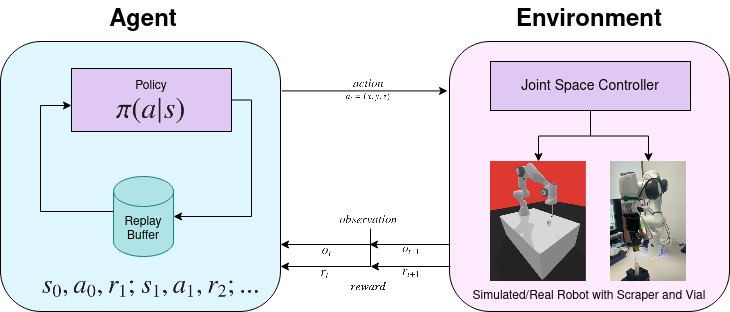}
    \caption{The overall block diagram of autonomous robotic scraping. Our method consists of an RL policy and a robotic controller. We map the action (cartesian pose) to joint positions using inverse kinematics to achieve the target goal. The controller uses force/torque feedback at the end effector and a joint space controller for motion planning.
    }
    \label{fig:overall_block_diagram} 
\end{figure*}

The goal of this work is to introduce and evaluate the performance of learning-based autonomous robotic laboratory skill acquisition, particularly for the task of scraping a vial sample.
This task is commonly performed in laboratory environments and is fundamental for successful autonomous powder diffraction~\cite{Lunt2024}, ~\cite{Szymanski2023}. 
This is particularly crucial as synthesising molecules is costly~\cite{Pecharsky2003}.
As illustrated in Fig.~\ref{fig:overall_block_diagram}, a real or simulated robotic manipulator uses a scraper that is attached to the robot's end effector and a vial is placed within its workspace.
The robot has to learn to insert the tool while scraping the vial walls.
Our proposed method relies on force/torque information at the robot’s end-effector which is essential  to realise contact-rich manipulation tasks; scraping glassware requires stringent force control which can be captured through sensory feedback. 
Force feedback is particularly intriguing as monitoring fine powder manipulation through perception is challenging due to the ubiquitous nature of the task: transparent media e.g. lab glassware suffer from specular reflections, and adding white, or transparent crystals makes it even more problematic.
In addition, given the contact-rich nature of the task, we believe this can only be captured through force feedback rather than visual demonstrations or feedback as different materials exhibit different properties when exposed to environmental factors.



\subsection{Problem Formulation}

\begin{figure}
    \centering
    \includegraphics[width=0.45\textwidth]{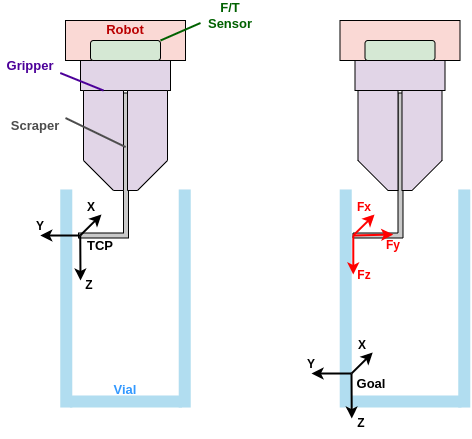}
    \caption{An illustration of the robotic scraping task. The robot inserts the scraper inside the vial. 
    The robot's goal is to translate along the z-axis while maintaining contact with the vial wall ($F_y \geq C_{threshold}$).}
    \label{fig:scraping_task}
\end{figure}

The task \emph{scrape} is defined as the following: the goal is to reach the target position at the bottom of the vial, while maintaining contact with the vial wall for material removal. 
The end-effector commands and state readings are computed with respect to the tool centre point (TCP) at the tip of the scraper.
The task is successful if the distance between the target position and the end-effector is smaller within the goal tolerance, and contact between the scraper and the vial is maintained when the scraper has been inserted inside.
The overall problem is illustrated in Fig.~\ref{fig:scraping_task}.

For modeling the underlying decision-making problem of an agent scraping powder from the vial walls, we adopt the finite-horizon discounted Markov decision process (MDP) formalism~\cite{Bellman1957},~\cite{Puterman1994}. It can be defined as a tuple 
$M = (S, A, P, r, \gamma, T)$, consisting of 
a continuous state space $S$, a continuous action space $A$, 
a transition function $P(s_{t+1}|s_t, a_t)$ defining the probability of the environment transitioning to state $s_{t+1}$ after the agent takes action $a_t$ in state $s_t$,
a reward function $r(s_t,a_t)$ specifying the scalar reward the agent receives after taking action $a_t$ in state $s_t$, a discount factor $\gamma \in [0, 1)$, and a horizon $H$ specifying the number of steps per episode.
The actions are drawn from a stochastic policy which is a state-conditioned probability distribution over actions $\pi(a_t|s_t)$.
The goal of the agent is to find the optimal policy $\pi^*$, which maximises the expected sum of discounted future rewards $ \mathbb{E}[\sum_{i=0}^T \gamma^{i} r_{t+i}]$ when the agent selects actions in each state according to $\pi^*$.

The details about the agent’s actions and state space (observations), as well as the formulation of our reward function for autonomous robotic sample scraping are outlined next.

\subsection{Agent Description}
As robotic scraping is a continuous control task, where sample efficiency is desired and we are dealing with both continuous actions and state spaces, we opted for an off-policy actor-critic algorithm.
Precisely, our method uses a state-of-the-art method for continuous control: the truncated quantile critic (TQC)~\cite{Kuznetsov2020} which builds on soft-actor critic (SAC)~\cite{Haarnoja2018}, twin delayed deep deterministic policy gradient (TD3)~\cite{Fujimoto2018} and quantile regression deep q-network (QR-DQN)~\cite{Dabney2018}.
TQC takes advantage of quantile regression to predict a distribution for the value function. 
Similar to TD3, it truncates the quantiles predicted by different networks.
We also compare this method in the simulated environment to vanilla SAC. 
The agent also made use of hindsight experience replay (HER)~\cite{Andrychowicz17}, a technique that increases sample efficiency by replacing goals are by the achieved goals and in turn updating the policy with previous states, actions, and rewards stored in a buffer.

\subsection{Model Architecture}
The model architecture was implemented using the stable baselines 3 software package~\cite{Raffin2021} and the hyperparameters chosen are based on the optimised ones in the RL baselines zoo~\cite{Raffin20} for goal-conditioned DRL.
Given that the observation space contains the position of the end-effector, we chose a multiple input policy network which concatenates all states into one vector that is passed to the policy. 
The model hyperparameters are given in TABLE~\ref{table:hypeparams}.

\begin{table}[b]
\begin{center}
\caption{Hyperparameters used for the DRL algorithm}

\begin{tabularx}{0.347\textwidth}{c|c}
 \hline
\textbf{Hyperparameter} & \textbf{Selected value} \\
 \hline
 Network type & Multiple Input Policy \\
Hidden layers & 3 \\ 
Hidden layer neurons & 512 \\ 
Activation hidden layers & ReLU \\ 
Learning rate & $1e^{-3}$ \\
Replay buffer size &  10k (500 real robot)\\
Batch size & 2048 (64 real robot) \\
$\Delta$  & 0.05cm (real robot only) \\

\end{tabularx}
\label{table:hypeparams}
\end{center}

\end{table}







\subsection{Observations}
The observations used for the scraping task are proprioceptive and force feedback. 
The proprioceptive observation gives the cartesian position of the end-effector ($x$, $y$, $z$). 
The force feedback describes the force/torque feedback at the end-effector ($F_X$, $F_Y$, $F_Z$).

\subsection{Actions}
The actions used in our experiments comprise the cartesian pose where $a = (\Delta x, \Delta y, \Delta z)$, which is the change in end-effector pose needed to arrive at the desired goal.
The $\Delta$ was limited such that the robot does not experience sharp changes and abrupt movements.
In practice, we converted the cartesian pose to joint angles of the seven DoF robotic manipulator via inverse kinematics to carry out the motion planning in joint space. 
We also limit these values (by scaling the output of the agent) to keep the angles within the joint limits and hence within a feasible range for the real robot and also to assure a more stable training process.

\subsection{Reward}

Our reward function for the simulated environment consists of the following components:

\begin{equation}
    R_{simulated\_task} = R_{goal} + w_0R_{contact} + w_1R_{vial}
    \label{eqn:reward_sim}
\end{equation}

At each time step, we calculate the shaping reward $R_{goal}$ as the L2 norm (Euclidean distance) between the target goal and current observation, defined as:

\begin{equation}
   R_{goal} = || goal_{desired} - goal_{current} ||_2
\end{equation}

$R_{goal}$ is motivated by observations of how chemists scrape samples as the manoeuvre is more often done vertically rather than horizontally.
From a technical viewpoint, this also makes it more feasible as a robotic task to maintain contact with the vial wall, as the translation is more similar to a plane rather than a curved surface.

The simulated agent's actions are constrained by two additional reward components that punish undesired behaviour: $R_{contact}$ and $R_{vial}$.
$R_{contact}$ is set to \textit{-1} if the horizontal part of the scraper does not touch the simulated vial represented.
In simulation, collision detection is used to verify where these two bodies are in contact, whereas for the real robot force/torque feedback at the end effector was measured directly.
$R_{vial}$ is set to \textit{-0.1} if the centre of the simulated vial is moved by 2cm. 
We explored different weightings of both rewards ($w_0$, $w_1$).  
The reward function for the real robotic scraping is similar to Equation~\ref{eqn:reward_sim} but without the penalty for moving the vial:

\begin{equation}
    R_{real\_task} = R_{goal} + R_{contact}
    \label{eqn:reward_real}
\end{equation}

In practice, vials are placed in racks; hence, the motion of the vial is already constrained, and we did not observe large displacements.

\section{Experimental Evaluation}
We evaluated our autonomous robotic scraper in simulation and on a real robotic platform.
Our experiments aim at answering the following research questions (RQ):
(RQ1) can we learn a policy that enables a robotic manipulator to scrape the contents of a vial? and 
(RQ2) given that the goal of scraping is to maximise the removal of powder from the vial walls, how does the policy deployed on the real robot perform?
This first experiment (addressing RQ1) aims to compare the performance of different off-policy model-free RL algorithms to learn a scraping policy for our task. 
The criteria for this comparison is the average model success rate and the mean return attained by different methods. 
These are both significant for the selection of the specific model that will be deployed, as we aim at minimising real-world training and inference times and the risk of failure. 
We trained directly in the real world as contact-rich tasks are hard to model in simulated environments, in addition to the non-trivial challenge of modelling powder and crystals in a simulated vial.
The second experiment (addressing RQ2) aims to illustrate that learning a good policy is not sufficient, but it is also important to visually demonstrate the removal of powder from the vial walls for task success.
Our demonstration is vital to understand failure cases that might hinder task completion, while giving us a better understanding of current shortcomings of the learning algorithms and the scraper (tool) so that we can optimise it for both our robotic manipulator and human scientists.

\subsection{Simulated Experiment}
\subsubsection{Experimental Setup}
Our simulated environment, illustrated in Fig.~\ref{fig:overall_block_diagram}, builds on panda-gym~\cite{Gallouedec2021}, which is a set of DRL environments for the Franka Emika Panda robot integrated with OpenAI gym.
The panda-gym benchmark does not yet have insertion tasks and we believe that our task provides an interesting novel environment here, especially for in-tube manipulation.
Moreover, the task of scraping is even more challenging than insertion as contact has to be maintained; hence our environment is new  as it introduces a constrained wiping task which is compelling to the robotics and machine learning communities alike.
In our task, the robot has 7 DoFs and it is equipped with a scraper attached to joint 7.
This was done to leverage quick experiments while focusing on the difficulty of contact-rich insertion and scraping rather than grasping.
The scraper consists of a white cylinder and a black box and was modelled based on a scraping tool that is commonly found in laboratory environments (Fig.~\ref{fig:scraping_workflow_images}).
In simulation, the vial used in a laboratory setting was modelled as a hollow tube.
All experiments were run using a machine equipped with an AMD Ryzen Threadripper 3970X 32 Core CPU and a single NVIDIA GeForce RTX 3090 Graphical Processing Unit (GPU).


\subsubsection{Experiment I - Learning a scraping policy}
\label{sssection:exp1}
\begin{figure}
    \centering
    \includegraphics[width=0.49\textwidth]{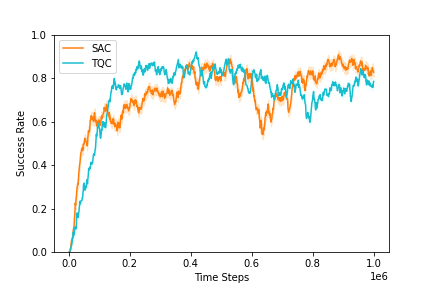}
    \caption{Evaluation results of learning a scraping policy using TQC and SAC in simulation across 5 random seeds.}
    \label{fig:exp1_results}
\end{figure}

In this experiment, the scraping task needs to be learned by only relying on the proprioceptive feedback and the contact force readings at the end-effector, which in simulation the collision detection between two bodies is used to monitor contact. 
The robot needs to learn a policy where the goal is to minimise the distance between the start and the goal pose, while maintaining contact with the inside of the vial wall.
During training, a target position is randomly generated within a pre-defined region on the bottom of the vial. 
The start pose is fixed and is set at the top of the vial.
The length of a trajectory is 50 timesteps, where at the end of each trajectory, the environment is reset and a new goal is randomly generated. 
All these trajectories are stored in a common replay buffer. 
The results are the success rate evaluated every 1000 test episodes over the course of learning. 
We explore two baselines of the results for the off-policy RL algorithms (TQC, SAC) used with HER. 
We report the evaluation performance of the best performing hyper-parameters for all algorithms (based on training performance).
We run all experiments with 5 seeds and report both mean and standard error (denoted by the shaded area on the plot) in Fig.~\ref{fig:exp1_results}.

While we explored the possibility of the robot scraping the full circular vial, we decided that this is not the best solution as human chemists scrape vertically and with the other arm they rotate the vial.
Furthermore, this process proved to be more robust for this task when deployed on the robot 


\subsubsection{Experiment II - Adding a curriculum}
\begin{figure}[!tbp]
  \centering
  \subfloat[]{\includegraphics[width=0.49\textwidth]{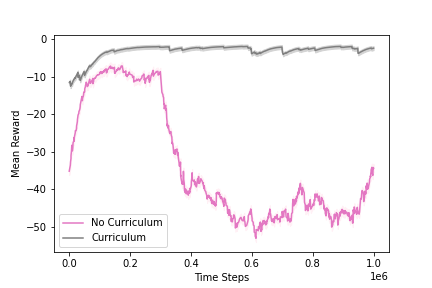}\label{fig:rewardexp1}}
  \hfill
  \subfloat[]{\includegraphics[width=0.49\textwidth]{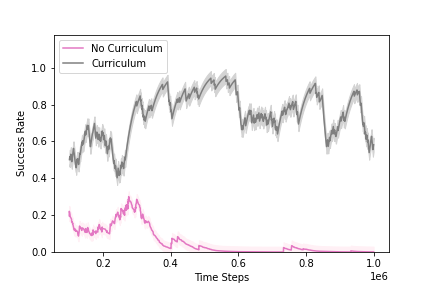}\label{fig:r_scraping}}
    \caption{Evaluation results of learning a scraping policy, using TQC, with a curriculum in simulation across 5 random seeds. We can see that using curriculum learning has higher success chances with increasing task difficulty, while using no curriculum learning is unable to perform well.}
\label{fig:exp2_results}
\end{figure}

This experiment focused on increasing the challenge of the scraping task by changing the starting pose to outside the vial, such that the task now covers both insertion and scraping.
The rest of the experiment is set up similar to Experiment I.
In this experiment we explore curriculum learning, where the key idea is to learn first on simple examples before moving to more difficult problems. 
In order to learn the target task, the agent first trains on a curriculum that consists of a sequence of intermediate tasks with increasing difficulty.
Our curriculum consists of only a single intermediate task, where the starting state is closer to the goal (at the top of the vial wall). 
This approach makes use of transfer learning methods to effectively extract and pass on reusable knowledge acquired in one task to the next. 
In our experiment, we achieve this by transferring the policies learned.
The results are the success rate evaluated every 1000 test episodes over the course of learning. 
We report the evaluation performance of the best performing hyper-parameters for TQC (based on training performance).
We run all experiments with 5 seeds and report both mean and standard error (denoted by the shaded area on the plot) in Fig.~\ref{fig:exp2_results}.
The results obtained demonstrate that using a curriculum with an intermediate task is necessary in order to achieve high success rates and generalise across different, more challenging start poses.

\subsection{Real Robot Experiment: An Autonomous Sample Scraping Case Study}
Our developed method was evaluated in the context of an autonomous laboratory workflow to demonstrate its potential to the challenging scraping problem.
This aim was to autonomously scrape powder off the walls of a given sample vial in the real world. 

\subsubsection{Experimental Setup}
The real setup comprised the Franka Emika Panda robot, equipped with a Robotiq 2F-85 parallel gripper and force torque sensor (FT 300-S). 
We trained our policy on hardware using a plastic vial, and then tested the model on a glass vial.
This was done to minimise potential glass breakage during training exploration.
Glass is used in laboratories as it has a high chemical compatibility with organic solvents and hence using plastic vials would limit the chemistry experiments. Our experimental setup is shown in Fig.~\ref{fig:scraping_workflow_images}.
For the real-world experiments, we used the TQC model.

\subsubsection{Case Study - Scraping powder}
The overall scraping task is demonstrated in the video provided in the supplementary material.
To monitor the progress of the scraping task a number of parameters were defined. 
We used food grade material (flour) since experiments were carried out on an open bench.
The vial was divided into $N_{reg}$ regions, since our learnt policy was constrained to an area of the vial equivalent to $\pi/12$ radians to avoid reaching the joint limit on the end effector. 

\begin{figure}[!tbp]
    \centering
    \subfloat[]{\includegraphics[width=0.4\textwidth]{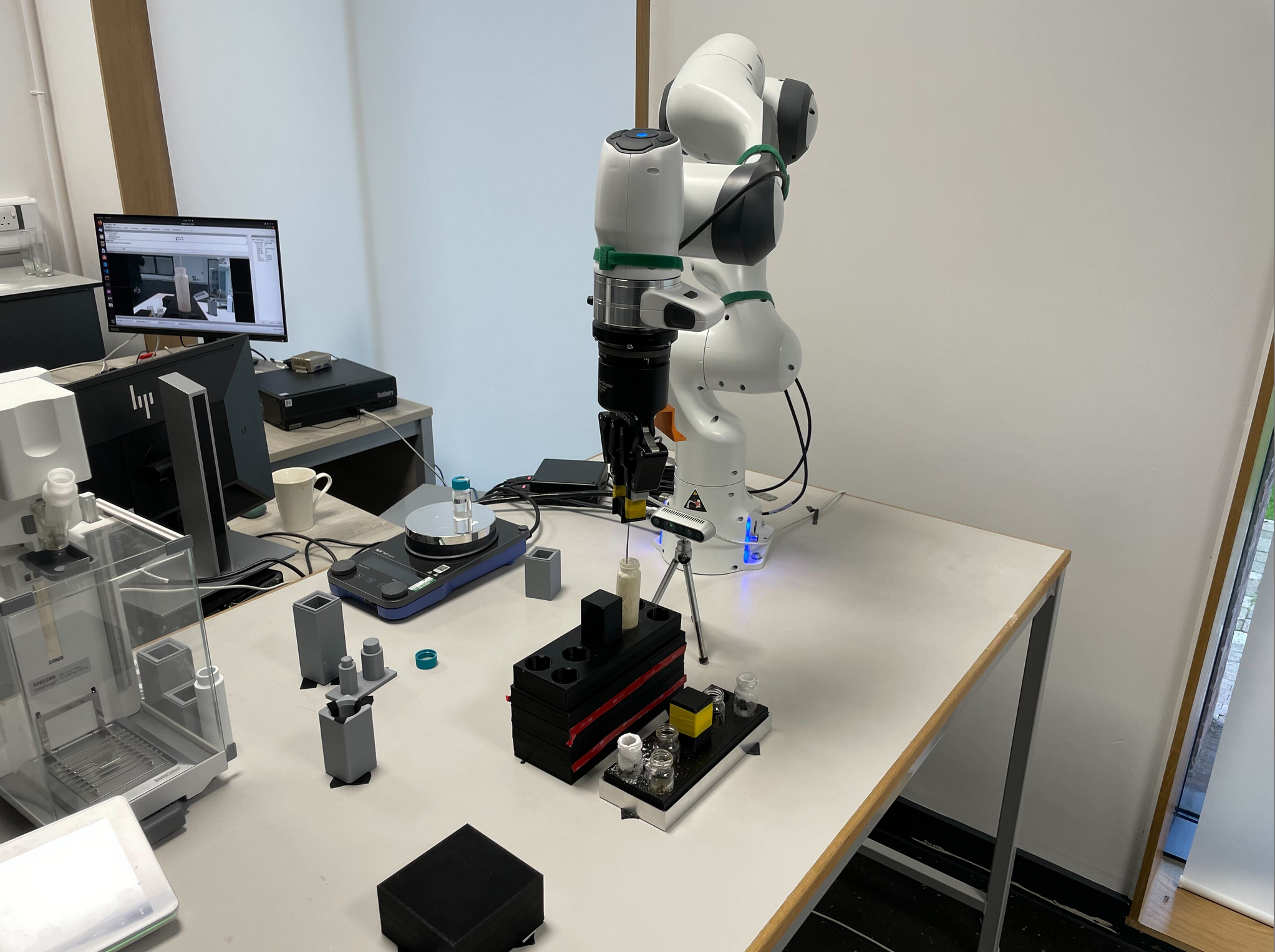}}
    \hfill
    \subfloat[]{\includegraphics[width=0.4\textwidth, trim={12cm 6cm 12cm 12cm},clip]{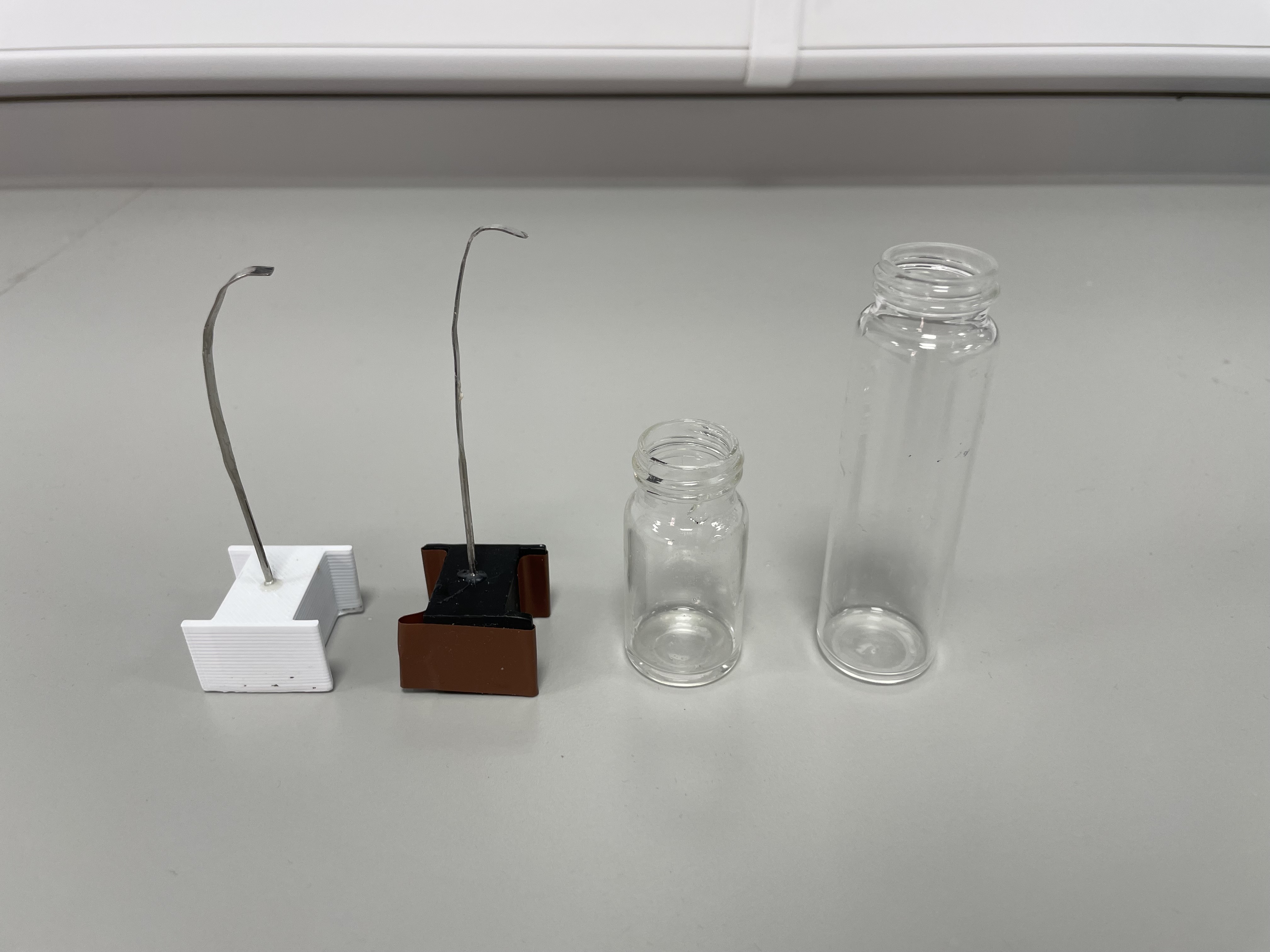}}
    \caption{An overview of (a) the robotic scraping setup (a complete video is provided as supplementary material) and (b) scrapers and vials used.}
    \label{fig:scraping_workflow_images}
\end{figure}

To complete the task, the robot takes the following steps. 
First, it picks up the scraping tool and goes to the scraping start pose. 
The robot then follows the scraping policy, where it picks the best sequence of actions to achieve the target goal. 
The robot would repeat this step for a pre-defined duration that was optimised to maximise scraping within the region in $N_{reg}$; in our experiment, this was chosen to be 5 times. 
This is followed by the robot putting the scraper in its holder and rotating the vial by $\pi/12$ radians, in preparation for scraping the next region in $N_{reg}$.
Following this, the robot grasps the tool again. 
Once the robot has carried this out across all regions ($N_{reg}=24$ if each region is $\pi/12$), the task is considered to be complete. 
This task was carried out successfully with two different sized vials and two scrapers of different length, as illustrated in Fig.~\ref{fig:scraping_workflow_images}.

\begin{figure*}[!tbp]
  \centering
  \subfloat[]{\includegraphics[width=0.33\textwidth, trim={6cm 0 6cm 0},clip]{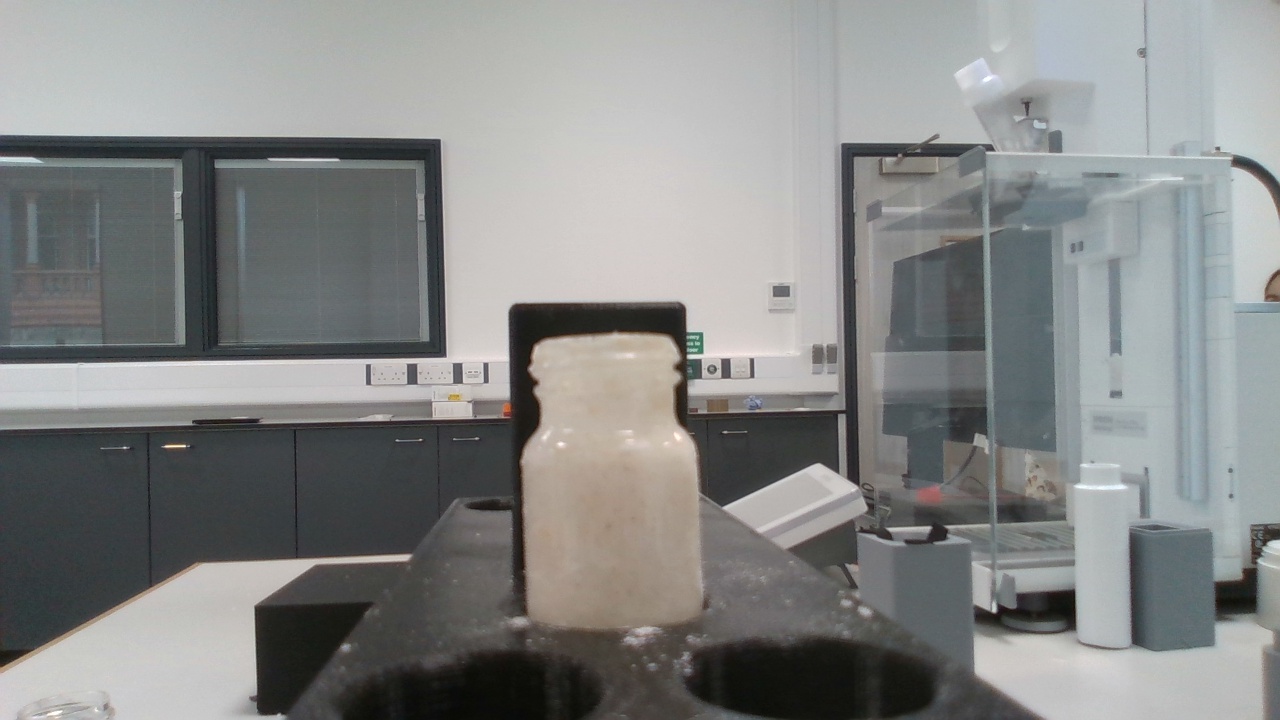}\label{fig:before}}
  \hfill
  \subfloat[]{\includegraphics[width=0.33\textwidth, trim={6cm 0 6cm 0},clip]{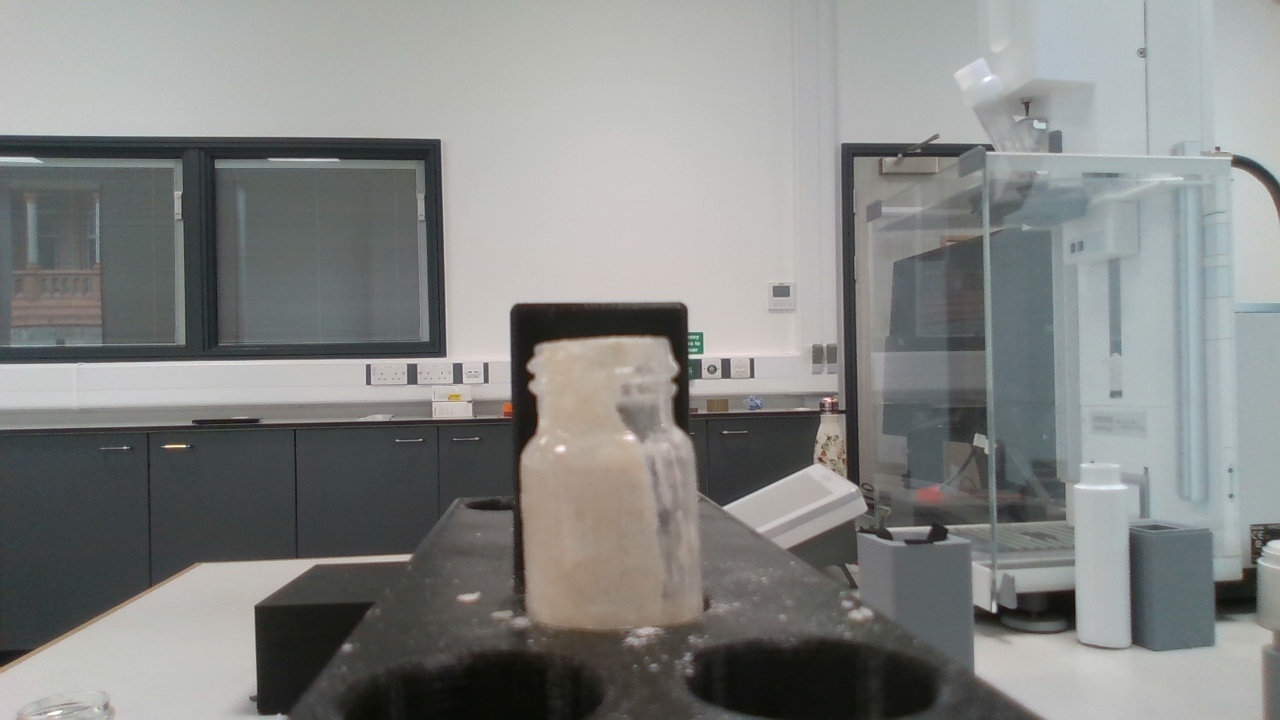}\label{fig:midway}}
  \subfloat[]{\includegraphics[width=0.33\textwidth, trim={6cm 0 6cm 0},clip]{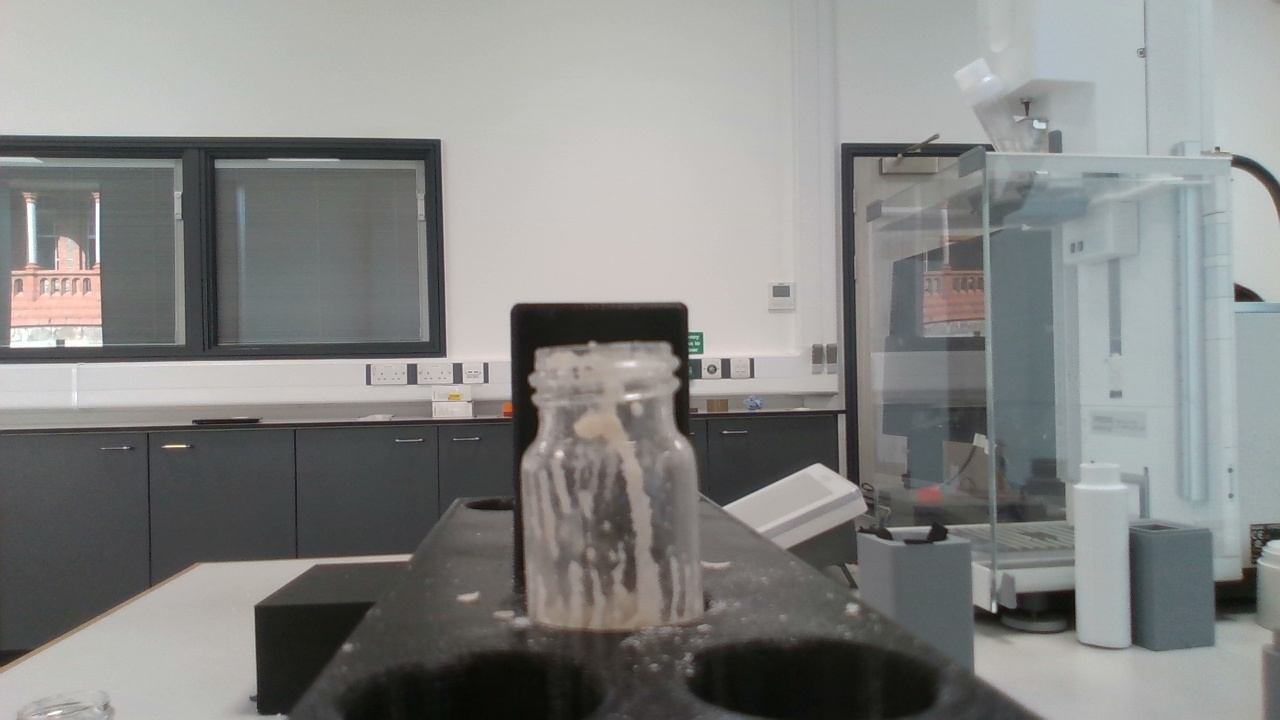}\label{fig:after}}  
  \caption{Qualitative results of the vials obtained from the case study showing a good performance of the proposed learning method, where (a) the initial vial condition, (b) midway through (powder partially removed) and (c) the powder scraped off.}
  \label{fig:vialfig}
\end{figure*}

In Fig.~\ref{fig:vialfig}, we demonstrate the results obtained from the autonomous robotic scraping.
Fig.~\ref{fig:before} demonstrates the vial at the start of the experiment with all of the powder still fixed inside the vial, whereas Fig.~\ref{fig:after} captures the vial contents after robot has scraped the inside of the vial.
The overall video is demonstrated \href{https://www.youtube.com/watch?v=0QQ3VPy0W6E}{here}.
In reality, scraping a fully-covered sample takes around 90 minutes; however, each region is scraped 5 times to ensure that all the powder has been scraped off the vial.
In addition, real samples from crystallisation workflows might only grow crystals on one side of the vial, and with an adequate perception method, our scraping policy could focus on solely that region. 
This would as a result speed up the scraping procedure.

\subsection{Failure Cases}
During the experimental trials, we analysed in-depth the failure cases.
An early failure case was when the actual robot pushed the inside of the vial that was outside of a rack and it toppled over.
In this instance, it was impossible for it to know that it had failed the task.
Hence, we made the decision of scraping vials when they are inside a rack, which is also a more realistic setting in a lab as it is rarely the case that vials are left outside of a rack on a bench.
We also introduced a threshold if the force $F_z$ is exceeded (20Nm) the robot should return to its start position, to minimise the chance of the robot pushing vials over and breakages.
We only set a threshold to $F_z$ as we did not notice the robot applying large forces for $F_x$ and $F_y$, particularly since the translation was predominantly in the z-direction.
Another similar example was when the scraper gets stuck in the vial and the robot could not generate enough force to reset and pull the scraper out.
In this case, the human needs to reset the robot manually.
This was the only non-recoverable failure case which required a human supervisor to reset the task.

\subsection{On Scraper Tool Choice}
The scraper used here is a tool that is commonly found in laboratories.
However, given that a robotic manipulator does not have the same dexterity as a human arm, the tool made the task more challenging due to the small surface contact that can be achieved. 
We also limited the change in motion during the scraping task to minimise high impact that could increase the risk of glass breakage.
As a result, this made the process slower but safer.
In the future, we will explore other tools such as brushes or even scrapers where the tool end is similar to a 2D cone to increase task efficiency.

\section{Conclusion}
In this work, we demonstrated that model-free reinforcement learning is a viable approach for learning laboratory skills such as autonomous robotic sample scraping. 
To learn the policy, we first created a simulated environment based on our real `robotic scientist' with a scraper and vial. 
This novel task challenges previous insertion benchmark environments as it requires a constrained powder wiping task during the insertion.
In the simulated environment we showed how distinct policies were learned for several off-policy RL approaches and that using curriculum learning is beneficial to generalise across more challenging configurations.
We then demonstrated that a data-driven model can be learned and deployed on a real robotic manipulator to successfully scrape the inside of a sample vial.
Our empirical results demonstrate that our proposed method enables the robot to complete the task across different vial sizes and tools.
Additionally, the results show the use of DRL enables us to complete tasks that have not been possible in chemistry workflows.
There are several future directions for extending and improving this approach.
On our immediate future plan, we want to explore how we can combine the visual input within the current reward.
Moreover, we want to look into bi-manual manipulation to speed up the scraping task where an arm is used to rotate the vial while the other one scrapes.



\bibliographystyle{ieeetr}
\bibliography{references}

\end{document}